\documentclass[wcp]{jmlr}
\renewcommand{\cite}{\citep}

\usepackage{amsmath}
\usepackage{amsthm}
\newtheorem{defn}{Definition}
\theoremstyle{plain}

\usepackage[utf8]{inputenc}
\usepackage{natbib}

\usepackage{wrapfig}

\makeatletter
\def\set@curr@file#1{\def\@curr@file{#1}} %temp workaround for 2019 latex release
\makeatother
\usepackage[load-configurations=version-1]{siunitx} % newer version

\jmlrworkshop{ACML 2020}
\jmlrpages{}

  \author{\Name{Dung {Nguyen}} \Email{dung.nguyen@deakin.edu.au}\\
  \Name{Svetha {Venkatesh}} \Email{svetha.venkatesh@deakin.edu.au}\\
  \Name{Phuoc {Nguyen}} \Email{phuoc.nguyen@deakin.edu.au}\\
  \Name{Truyen {Tran}} \Email{truyen.tran@deakin.edu.au}\\
  \addr Applied Artificial Intelligence Institute, Deakin University, Geelong, Australia
 }

\title[ToMAGA]{Theory of Mind with Guilt Aversion \\ Facilitates Cooperative Reinforcement Learning}
\begin{document}

\maketitle

\begin{abstract}
    Guilt aversion induces experience of a utility loss in people if they believe they have disappointed others, and this promotes cooperative behaviour in human. In psychological game theory, guilt aversion necessitates modelling of agents that have theory about what other agents think, also known as Theory of Mind (ToM). We aim to build a new kind of affective reinforcement learning agents, called Theory of Mind Agents with Guilt Aversion (ToMAGA), which are equipped with an ability to think about the wellbeing of others instead of just self-interest. To validate the agent design, we use a general-sum game known as Stag Hunt as a test bed. As standard reinforcement learning agents could learn suboptimal policies in social dilemmas like Stag Hunt, we propose to use belief-based guilt aversion as a reward shaping mechanism. We show that our belief-based guilt averse agents can efficiently learn cooperative behaviours in Stag Hunt Games.
\end{abstract}

\section{Introduction}
People in a group may be willing to give more and take less. This
may appear irrational from the individual perspective, but such behaviour
often enables the group to achieve higher returns than acting selfishly.
Therefore, in building artificial multi-agent systems, it is important
to construct social inductive biases about the reasoning of other
agents - also known as the Theory of Mind (ToM)~\cite{rabinowitz2018machine,shum2019theory}.
Theory of mind enables individuals to cooperate and this often results
in optimal group rewards \cite{shum2019theory,takagishi2010theory}.

A mechanism to encourage social cooperation is maintaining \emph{fair}
outcomes for members of the group, and agents who do so are termed
`inequity averse' \cite{hughes2018inequity}. Other mechanisms stem
from \emph{guilt} \cite{haidt2012righteous}, requiring one to put
themselves in the others' shoes \cite{chang2011triangulating,morey2012neural}.
To be \emph{guilt averse, the agent needs higher-order ToM - i.e.
}be able to estimate\emph{ }what others will do (0-order ToM), and
what others believe the agent itself will do (1-order ToM) \cite{albrecht2018autonomous}\emph{.
}Inequity aversion, on the other hand, is conceptually different from
guilt aversion \cite{nihonsugi2015selective} and does not require
theory of mind. We focus on the computational mechanisms to control
the interplay between the greedy tendencies of an individual and the
inferred needs of others in a reinforcement learning (RL) setting.
In \cite{moniz2017social,rosenstock2018s}, authors analysed the evolutionary
dynamics of agents with guilt, but did not include theory of mind.
There has been early work to integrate theory of mind and guilt aversion
in a psychological game setting \cite{battigalli2007guilt}. The first
work to examine social dilemmas in a deep reinforcement learning setting
is \cite{hughes2018inequity,peysakhovich2018prosocial} in which the
authors incorporate knowledge from behavioural game theory when training
the agents. However, guilt aversion, which plays a central role in
moral decisions \cite{haidt2012righteous} has not been considered.

Our paper addresses the open challenges of integrating theory of mind
and guilt aversion into Multi-Agent Reinforcement Learning (MARL)
\cite{littman1994markov} and studies the evolution of cooperation
in such agents in self-play settings. We name the agent ToMAGA, which
stands for \emph{Theory of Mind Agent with Guilt Aversion}. In our
agents, learning is driven by not only material rewards but also psychological
loss due to the feeling of guilt if an agent believes that it has
harmed others. Our computational model of theory of mind extends the
work of \cite{de2013much} to build agents with beliefs about cooperative
behaviours rather than just primitive actions. Our reinforcement learning
agent uses a value function to make sequential decisions. At each
learning step, after observing the other agents' actions, the agent
updates its beliefs about the other agents, including what they might
think about it. Then it computes psychological rewards using a guilt
averse model, followed by an update of the value function. In other
words, this implements a reward shaping strategy, where the additional
reward is from the intrinsic social motivation of being fair to others.
In reinforcement learning, reward shaping helps to guide the exploration
and increase the convergence speed of the algorithm. Different from
\cite{devlin2011theoretical} in which the reward shaping function
was defined over the state space, our reward shaping function, on
the other hand, is defined over actions space. To help understand
how this reward shaping is effective in social dilemmas, we construct
a theoretical argument to show that guilt aversion implemented as
reward shaping can change the Stag Hunt game from having two pure
Nash equilibria into having one pure Nash equilibrium that is Pareto
efficient. In addition, our agents are able to cooperate in the grid-world
Stag Hunt Games, in which the rewards given to each agent depend on
the sequence of actions (at the policy level), not just on one action
like in matrix-form games. We build several environments, both in
a one-step decision game and in a multi-step grid-world. Our extensive
suite of experiments demonstrates that modelling guilt with explicit
theory of mind helps reinforcement learning agents to cooperate better
than those without theory of mind, encouraging faster learning towards
cooperative behaviours. At last, we demonstrate the efficiency of
our reward shaping mechanism on more complex rewards structure and
action space environments. We also demonstrate that the mechanism
can handle the case in which there are more than two agents.

Our contribution is to design and test a framework that brings the
psychological concept of guilt aversion into multi-agent reinforcement
learning, and in effect it connects social psychology, psychological
game theory \cite{geanakoplos1989psychological}, multi-agent systems
and reinforcement learning. For the first time, we explore and establish
a computational model for embedding guilt aversion coupled with theory
of mind on reinforcement learning framework and study it in the extended
Markov Games.

\section{Preliminaries}
\subsection{Two-player Markov Games}

A two-player fully observable Markov Game is a tuple $\left\langle \mathcal{N},\mathcal{S},\mathcal{A},\mathcal{R},\mathcal{P}\right\rangle $,
where $\mathcal{N}=\{1,2\}$ denotes the set of two players, $\mathcal{S}$
is the state space, $\mathcal{A=}\mathcal{A}_{1}\times\mathcal{A}_{2}$
is the joint action space, $\mathcal{R=}\mathcal{R}_{1}\times\mathcal{R}_{2}$
is the reward space with $\mathcal{R}_{1},\mathcal{R}_{2}:\mathcal{S}\times\mathcal{A}\mapsto\mathbb{R}$,
$\mathcal{P}$ is the transition function $\mathcal{P}:\mathcal{S}\times\mathcal{A}\mapsto\Delta(\mathcal{S})$,
where $\Delta(\mathcal{S})$ denotes the probability distribution
over $\mathcal{S}$. Each agent $i$ takes an action $a_{i}\in\mathcal{A}_{i}$
based on its policy $\pi_{i}:\mathcal{S}\mapsto\mathcal{A}_{i}$.
Denote by $\Pi_{i}$ the set of all policies available to the player
$i$. The set of joint policies is $\Pi=\Pi_{1}\times\Pi_{2}$. 
\begin{defn}
A joint policy $\pi=(\pi_{1},\pi_{2})\in\Pi$, denoted as $\pi^{C}=(\pi_{1}^{C},\pi_{2}^{C})$,
is a \emph{cooperative joint policy }iff

\begin{align*}
\pi & =\text{argmax}_{\pi_{1}\in\Pi_{1},\pi_{2}\in\Pi_{2}}\mathbb{E}_{a_{1}\sim\pi_{1},a_{2}\sim\pi_{2},s_{t+1}\sim\mathcal{P}}\left[R\right],\,\text{for}\\
R & =\sum_{t=0}^{\infty}\gamma^{t}\left[r_{1}(a_{1},a_{2},s_{t})+r_{2}(a_{1},a_{2},s_{t})\right]
\end{align*}
If two agents follow a \emph{cooperative joint policy} $\pi^{C}$,
we say that two agents have \emph{cooperative behaviours.} We denote
$\Pi^{C}$ as a set of \emph{cooperative joint policies}. 
\end{defn}

\begin{defn}
A policy $\pi_{i}$ is a \emph{cooperative policy} iff

\[
\exists j\in\mathcal{N}\backslash i,\pi_{j}\in\Pi_{j}:(\pi_{1},\pi_{2})\in\Pi^{C}.
\]
We denote $\Pi_{i}^{C}$ as a set of cooperative policies. A policy
$\pi_{i}\in\Pi_{i}\backslash\Pi_{i}^{C}$ of agent $i$ is called
an \emph{uncooperative policy}. We denote $\Pi_{i}^{U}=\Pi_{i}\backslash\Pi_{i}^{C}$
as a set of uncooperative policies\emph{.} 

The definition says that if the agent $i$ follows policy $\pi_{i}$
and there exists at least one policy $\pi_{j}$ of agent $j$ that
their joint policy $\pi=(\pi_{1},\pi_{2})$ is a cooperative joint
policy, then the policy $\pi_{i}$ a cooperative policy.
\end{defn}

\subsection{The Stag Hunt Game \label{subsec:The-Stag-Hunt}}

\begin{table}
\begin{centering}
\begin{tabular}{|c|c|c|}
\hline 
 & $C$ & $U$\tabularnewline
\hline 
\hline 
$C$ & $h,h$ & $g,c$\tabularnewline
\hline 
$U$ & $c,g$ & $m,m$\tabularnewline
\hline 
\end{tabular}
\par\end{centering}
\caption{\label{tab:matrix_form_StagHunt} The structure of Stag Hunt $(h>c>m>g)$. }
\end{table}

Stag Hunt is a coordination game of two persons hunting together \cite{macy2002learning}.
If they hunt stag together, they can both obtain a large reward $h$.
However, one can choose to trap hare gaining a reward, sacrificing
the other's benefit. The reward matrix is shown in Table~\ref{tab:matrix_form_StagHunt}.
The game has two pure Nash equilibria: (1) both hunting stag, which
is Pareto optimal; (2) or both hunting hare. If one player thinks
the other will choose to hunt hare, her best response will be hunting
hare. This is because in the worst case, if hunting hare, the player
will receive a reward $m$. This amount is larger than $g$ which
is the worst case if she hunts stag. Therefore, both hunting hare
is the \emph{risk-dominant }equilibrium \cite{harsanyi1988general}.
Here, the dilemma is that the \emph{risk-dominant} Nash equilibrium
is not the Pareto optimal. There is one mixed Nash equilibrium but
its common outcome is not Pareto optimal. Because both will receive
the highest collective rewards when jointly hunting stag, both hunting
stag is a joint cooperative policy. Therefore, hunting stag is a cooperative
policy that is also Pareto efficient.

\section{Theory of Mind Agents with Guilt Aversion}
% %% LyX 2.3.5.2 created this file.  For more info, see http://www.lyx.org/.
% %% Do not edit unless you really know what you are doing.
% \documentclass[british]{jmlr}
% \usepackage[T1]{fontenc}
% \usepackage[latin9]{inputenc}
% \usepackage{amsmath}
% \usepackage{amsthm}
% \usepackage{amssymb}
% \usepackage{graphicx}

% \makeatletter

% %%%%%%%%%%%%%%%%%%%%%%%%%%%%%% LyX specific LaTeX commands.
% %% Because html converters don't know tabularnewline
% \providecommand{\tabularnewline}{\\}

% %%%%%%%%%%%%%%%%%%%%%%%%%%%%%% Textclass specific LaTeX commands.
% \theoremstyle{plain}
% \newtheorem*{assumption*}{\protect\assumptionname}

% %%%%%%%%%%%%%%%%%%%%%%%%%%%%%% User specified LaTeX commands.
% %\usepackage[linesnumbered,algoruled,boxed,lined]{algorithm2e}

% \makeatother

% \usepackage{babel}
% \providecommand{\assumptionname}{Assumption}

\begin{figure}[t]
    \begin{centering}
        \includegraphics[width=0.65\columnwidth]{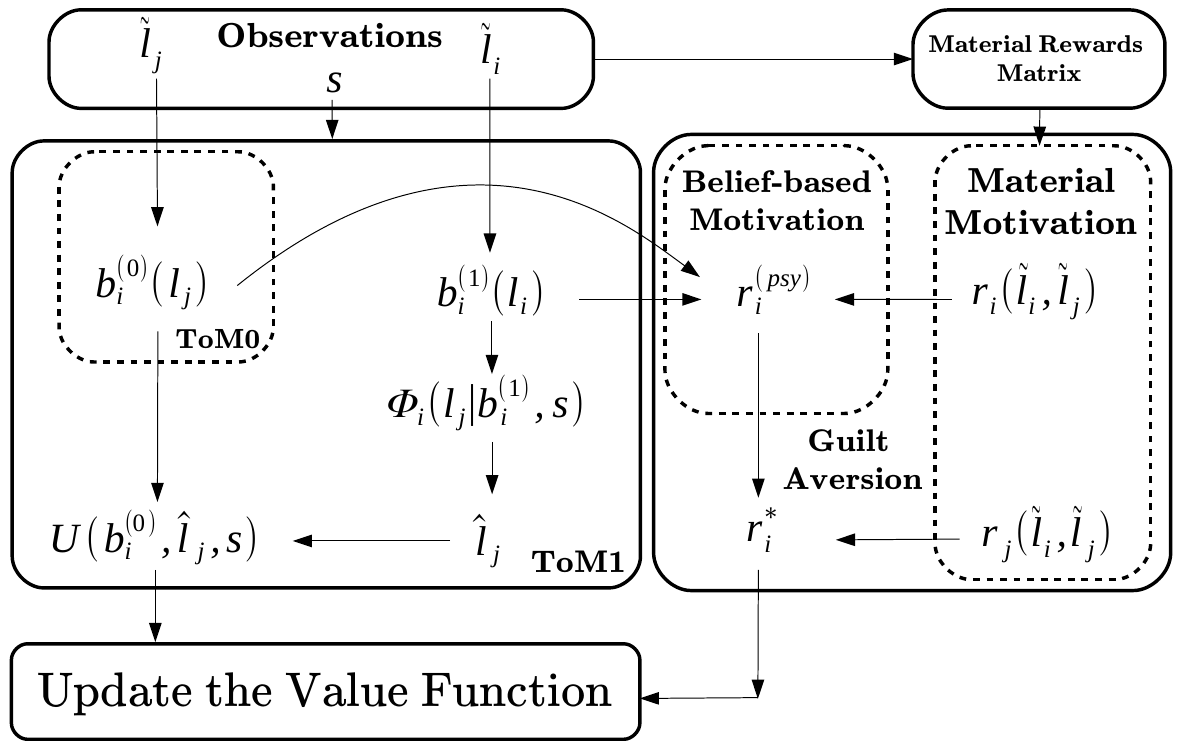}
        \par\end{centering}
    \caption{\label{fig:ToM2PsyIllu}The learning process in Theory of Mind Agents
    with Guilt Aversion (ToMAGAs).%\label{fig:Model-overview}
    }
\end{figure}

We present our agent model named \emph{Theory of Mind Agent with Guilt
Aversion} (ToMAGA). The internal working process of the agent is illustrated
in Fig.~\ref{fig:ToM2PsyIllu}. It has a ToM module that is augmented
with a guilt aversion (GA) component. We detail these parts as follows.

\subsection{Settings}

In our setting, an agent learns: (1) to predict whether the other
agent follows a cooperative policy or an uncooperative policy; and
(2) a cooperative policy.  These objectives are also a part of things
to learn in a fully observable games in the area of Multi-agent Learning
(MAL) \cite{shoham2007if}. During the $k^{th}$ iteration, agent
$i$ follows its policy $\pi_{i}^{(k)}$. For readability, we omit
the superscript $(k)$. At each time step $t$ of an iteration in
the game, two agents simultaneously take actions, hence, generating
a trajectory of \emph{experiences} $\tau=\left(s^{(t)},a_{1,}^{(t)}a_{2}^{(t)},r_{1}^{(t+1)},r_{2}^{(t+1)},s^{(t+1)}\right)_{t=0}^{T-1}$.
We make the following assumptions about the observations of agents
and the reward structure of the training environment

% \begin{assumption}
\paragraph{Assumption.}
\emph{(about the observations) In any iteration $k$, a policy $\pi_{i}$
for $i\in\mathcal{N}$ belongs to the set of joint cooperative policy\emph{
$\Pi_{i}^{C}$ }or the set of uncooperative policies\textup{ $\Pi_{i}^{U}$}.
Both agents can observe this information.}
% \end{assumption}

We denote by $l_{i}=C$ the event that at iteration $k$ the policy
of agent $i$ is a \emph{cooperative policy} $\pi_{i}\in\Pi_{i}^{C}$
and by $l_{i}=U$ the event that the policy of agent $i$ is an \emph{uncooperative
policy} $\pi_{i}\in\Pi_{i}^{U}$.
\paragraph{Assumption.}
% \begin{assumption}
\emph{(about the reward structure) After reaching the termination state of the game, the agents receive
material rewards $r_{1}^{(T)}\left(l_{i},l_{j}\right)$ and $r_{2}^{(T)}\left(l_{i},l_{j}\right)$.
The rewards follow the structure of the Stag Hunt Game described in
Table~\ref{tab:matrix_form_StagHunt}.}
% \end{assumption}

\subsection{First-order Theory of Mind (ToM1) Agent}

We construct ToM1 agents as in \cite{de2013much}. Agent $i$ maintains
two beliefs: (1) zero-order belief $b_{i}^{(0)}(l_{j})$ for $l_{j}\in\left\{ C,U\right\} $
which is a probability distribution over events that agent $j\ne i$
follows a cooperative\emph{ }or an\emph{ }uncooperative policy; and
(2) first-order belief $b_{i}^{(1)}(l_{i})$ for $l_{i}\in\left\{ C,U\right\} $,
which is a recursive belief, representing what agent $i$ thinks \emph{about
the belief of agent $j$'s belief (the probability distribution over
events that agent $i$ follows a cooperative or an uncooperative policy)}.
At the end of each iteration, agent $i$ observes a trajectory $\tau$
and the information about whether the executed policies were cooperative
or uncooperative, i.e. $\left\{ \tilde{l}_{i},\tilde{l}_{j}\right\} $.
Agent $i$ first predicts whether agent $j$ uses a cooperative or
an uncooperative policy. The prediction is based on the current first-order
belief $b_{i}^{(1)}(l_{i})$ as follows

\begin{align*}
\hat{l}_{j} & =\text{argmax}_{l_{j}\in\left\{ C,U\right\} }\varPhi_{ij}(l_{j})\,\,\text{where}\\
\varPhi_{ij}(l_{j}) & =\begin{array}{cc}
{\displaystyle \sum_{l_{i}\in\left\{ C,U\right\} }b_{i}^{(1)}(l_{i})\times r_{j}^{(T)}(l_{i},l_{j}),}\end{array}
\end{align*}
where $\varPhi_{ij}(l_{j})$ is the value function agent $i$ thinks
agent $j$ will have if agent $j$ greedily maximises its material
reward. Now, agent $i$ has two guesses about the agent $j$: the
zero-order belief $b_{i}^{(0)}(l_{j})$ and policy type $\hat{l}_{j}$.
 To combine these two pieces of information into the belief about
the action of agent $j$, called a belief integration function $BI(l_{j})$.
To do this, agent $i$ maintains and updates a confidence $c_{ij}\in\left[0,1\right]$
about its ToM1 as follows: 
\[
c_{ij}\leftarrow(1-\lambda)c_{ij}+\lambda\delta\left[l_{j}=\hat{l}_{j}\right]
\]
for learning rate $\lambda\in\left[0,1\right]$ and identity function
$\delta\left[\cdot\right]$. After updating the confidence, agent
$i$ then computes its belief integration function

\[
BI(l_{j})\leftarrow\left(1-c_{ij}\right)b_{i}^{(0)}(l_{j})+c_{ij}\delta\left[l_{j}=\hat{l}_{j}\right]
\]
for all $l_{j}\in\left\{ C,U\right\} $. Now the agent $i$ can update
its zero-order belief as

\[
b_{i}^{(0)}(l_{j})\leftarrow BI(l_{j}),
\]
for all $l_{j}\in\left\{ C,U\right\} $ and first-order belief as

\[
b_{i}^{(1)}(l_{i})\leftarrow\left(1-c_{ij}\right)b_{i}^{(1)}(l_{i})+c_{ij}\times\delta\left[l_{i}=\tilde{l}_{i}\right],
\]
for all $l_{i}\in\left\{ C,U\right\} $.

\subsection{Guilt Aversion (GA)}

The guilt averse agent $i$ will experience a utility loss if it
thinks it lets the other agent down. The utility loss is realised
through reward shaping. More concretely, once beliefs are updated,
the agent $i$ first computes an expected material value experienced
by the agent $j$:

\begin{equation}
\phi_{j}=\sum_{l_{i},l_{j}\in\left\{ C,U\right\} }b_{i}^{(0)}(l_{j})\times b_{i}^{(1)}(l_{i})\times r_{j}^{(T)}(l_{i},l_{j})\label{eq:expectation}
\end{equation}
where $r_{j}^{(T)}(l_{i},l_{j})$ is the material reward received
after the last time step $T$. In addition, the agent experiences
a psychological reward of ``feeling guilty'', caring about how much
it lets the other down, as \cite{battigalli2007guilt}:

\begin{equation}
r_{i}^{(psy)}(\tilde{l}_{i},\tilde{l}_{j})=-\theta_{ij}\max\left(0,\phi_{j}-r_{j}^{(T)}(\tilde{l}_{i},\tilde{l}_{j})\right)\label{eq:psy-reward}
\end{equation}
where guilt sensitivity $\theta_{ij}>0$. The reward is then shaped
as:

\begin{equation}
r_{i}^{*}=r_{i}^{(T)}(\tilde{l}_{i},\tilde{l}_{j})+r_{i}^{(psy)}(\tilde{l}_{i},\tilde{l}_{j}).\label{eq:reward-shaping}
\end{equation}

This computation is based on an assumption that a guilt averse agent
\emph{does not} know whether the other is guilt averse. 

\subsection{Update the Value Function}
\begin{algorithm} 
	\caption{ToMAGA $i$ } %in Grid-world Stag Hunt
	\label{algo:ToMAGA}
	\SetKwInOut{Input}{Input}   
	\SetKwInOut{Output}{Output}
	\Input{ $K$ is the number of iterations \newline $T_{max}$  is the maximum timesteps per iteration}%,  $\gamma$, $\alpha$, $\theta_{ij}$}   %$\epsilon$,
	\Output{ The policy $\pi_i $ }  % = \text{argmax}_{a_i} Q_i (s,a_{i})

	%Initialise $Q^{(0)}_{i}(s^{(0)},a_{i}^{(0)}) = 0$ for all states $s \in \mathcal{S}$ and actions $a_i \in \mathcal{A}_i$\; 
	%Initialise $Q^{(0)}_{i}(s^{(0)},a_{i}^{(0)}) = 0$ for all states and actions; 

	\For{ $k \leftarrow 0$ \KwTo $K-1$ } 
	{
		
		Reset $\tau^{(k)}$\;
		\For{ $t\leftarrow 0$ \KwTo $T_{max}$}{
			%\eIf{$k < K_{explore}$}{
			%	Takes action $\tilde{a}^{(t)}_{i} $ based on $\epsilon$-greedy;
			%}{
			Takes action $\tilde{a}^{(t)}_{i}$ based on the value function; \\ %= \text{argmax}_{a_i} Q_i(a_{i}, s^{(t)}) $ 
				%Takes action \\
				%\begin{center}
				%\centering $\tilde{a}^{(t)}_{i} = \text{argmax}_{a_i} Q_i(a_{i}, s^{(t)}) $\;
				%\end{center}							
			%}
			Add the new experience to $\tau$\;
			\If{$s_t$ {\normalfont is the} termination state}{
				\textbf{break}\;
			}

			%$\tau = \tau \cup (a_{1(t),}a_{2(t)},r_{1(t+1)},r_{2(t+1)},s_{t+1})$ 
			
		}
		Get information about policies $\left\{ \tilde{l}_{i},\tilde{l}_{j}\right\} $\;

		Update beliefs $b_{i}^{(0)}(l_{j})$ and  $b_{i}^{(1)}(l_{i})$ \;

		%Get information about policies 
		%Get $\left\{ \tilde{l}_{i},\tilde{l}_{j}\right\} $ and update $b_{i}^{(0)}(l_{j})$ and  $b_{i}^{(1)}(l_{i})$\;

		Compute psychological reward $r_{i}^{(psy)}(\tilde{l}_{i},\tilde{l}_{j})$\;

		%Update the value function  $Q_i$\;

		\ForAll{{\normalfont experiences in} $\tau$}
		{
			Update the value function by using $r^*_i$ in Eq. 3;%$Q_i(s^{(t)},a^{(t)}_{i})$\;
		}
	}
\end{algorithm}
Given the shaped reward in Eq.~(\ref{eq:reward-shaping}), the reinforcement
learning agent learns by updating the value function as follows.

\paragraph{Matrix-form Stag Hunt}
Because the size of the state space $|\mathcal{S}|=1$, the strategy
of agent $i$ reduces to select action $a_{i}$, and agent $i$ updates
its value function based on temporal difference algorithm TD$(1)$:

\begin{align*}
V_{i}(\tilde{a}_{i}) & \leftarrow V_{i}(\tilde{a}_{i})+\alpha\Delta_{i},\,\,\text{where}\\
\Delta_{i} & =r_{i}^{*}+\gamma\max_{a_{i}}{\displaystyle \sum_{a_{j}}b_{i}^{(0)}(a_{i})r_{i}(a_{i},a_{j})}-V_{i}(\tilde{a}_{i})
\end{align*}

\paragraph{General Stag Hunt with Deep Reinforcement Learning}
In the general Stag Hunt games, we parameterise the value function
and policy by deep neural networks trained by the Proximal Policy
Optimization (PPO) \cite{schulman2017proximal}. The training algorithm
is shown in Algorithm~\ref{algo:ToMAGA}.

\subsection{Theoretical Analysis}

We now show that guilt aversion implemented as reward shaping can
change the Stag Hunt game from having two pure Nash equilibria into
having one pure Nash equilibrium that is Pareto efficient. We recall
that ToMAGAs play the game with a new pay-off matrix as shown in Table~\ref{tab:matrix_form_psyStagHunt}.
We then establish the following observations.

\begin{table}
\begin{centering}
\begin{tabular}{|c|c|c|}
\hline 
 & $C$ & $U$\tabularnewline
\hline 
\hline 
$C$ & $h+r^{(psy)},h+r^{(psy)}$ & $g+r^{(psy)},c+r^{(psy)}$\tabularnewline
\hline 
$U$ & $c+r^{(psy)},g+r^{(psy)}$ & $m+r^{(psy)},m+r^{(psy)}$\tabularnewline
\hline 
\end{tabular}
\par\end{centering}
\caption{The reward structure of Stag Hunt games after having psychological
reward factor defined in Eq.~\ref{eq:psy-reward}. \label{tab:matrix_form_psyStagHunt}}
\end{table}

\paragraph{Observations:}

\emph{ }\textbf{(1)}\emph{ If there exists a sequence of trajectories
leading to $\phi_{j}>m$ and $\theta_{ij}>\frac{m-g}{\text{min}(\phi_{j},c)-m}$
with $i,j\in\{1,2\},i\neq j$, this game will have only one pure Nash
equilibrium, in which both players choose to cooperate $(C,C)$; and
}\textbf{(2)}\emph{ ToMAGA with higher guilt sensitivity $\theta_{ij}$
will have a higher chance of converging to this pure Nash equilibrium
in self-play setting.}
\begin{proof}
This game will have only one pure Nash equilibrium (NE), in which
both players choose to cooperate $(C,C)$, when two conditions hold:
\begin{align*}
(\text{C1})\,\,h-\theta_{ij}\text{max}(0,\phi_{j}-h) & >c-\theta_{ij}\text{max}(0,\phi_{j}-g)\\
(\text{C2})\:\,\,g-\theta_{ij}\text{max}(0,\phi_{j}-c) & >m-\theta_{ij}\text{max}(0,\phi_{j}-m)
\end{align*}
for $h>c>m>g$, the sensitivity $\theta_{ij}>0$, and the expected
material value experienced by other agent $\phi_{j}\in\left[g,h\right]$
described in Eq. \ref{eq:expectation}. (C1) holds within the structure
of the Stag Hunt game. When $\phi_{j}\in\left(c,h\right]$, (C2) is
satisfied iff $\theta_{ij}>\frac{m-g}{c-m}$. When $\phi_{j}\in\left(m,c\right]$,
(C2) is satisfied iff $\theta_{ij}>\frac{m-g}{\phi_{j}-m}$. Therefore,
the first observation is proved. To prove the second observation,
we consider the case when $\phi_{j}\in\left(m,c\right]$, the condition
$\theta_{ij}>\frac{m-g}{\phi_{j}-m}\Leftrightarrow\phi_{j}>\left(m+\frac{m-g}{\theta_{ij}}\right)\triangleq f(\theta_{ij})$
implies $\phi_{j}\in\left(f(\theta_{ij}),c\right]$. Because $f(\theta_{ij})$
is a decreasing function, the chance of $\phi_{j}$ belongs to $\left(f(\theta_{ij}),c\right]$
is increasing when $\theta_{ij}$ is increasing, i.e. the second observation
is proved. \textbf{\emph{}}
\end{proof}
By introducing the psychological rewards, we increase the probability
of changing the game from two Nash equilibria to one Nash equilibrium,
which intuitively helps the reinforcement learning algorithm converge
to Pareto efficient Nash equilibrium. In other words, the higher guilt
sensitivity the agents have, the more chance that they will converge
to the cooperative behaviour. During the exploration, both agents
need to obtain higher beliefs about the event that other will choose
a cooperative policy. If both agents believe that the expectation
of other are higher than the outcome that players received when they
are at risk-dominant Nash equilibrium, i.e. $\phi_{j}>m$, then higher
$\theta_{ij}$ will lead to higher chance to converge to NEs. However,
initially, if both agents believe that the expectation of other are
equal or lower than the inefficient outcome, i.e. $\phi_{j}\leq m$,
the agents \emph{need to} increase this expectation during the training
process.

\section{Experiments}
%% LyX 2.3.5.2 created this file.  For more info, see http://www.lyx.org/.
%% Do not edit unless you really know what you are doing.
% \documentclass[british]{jmlr}
% \usepackage[T1]{fontenc}
% \usepackage[latin9]{inputenc}
% \usepackage{wrapfig}
% \usepackage{amstext}
% \usepackage{graphicx}
% \usepackage{babel}
% \begin{document}
\begin{figure*}
\begin{centering}
\includegraphics[width=0.4\textwidth]{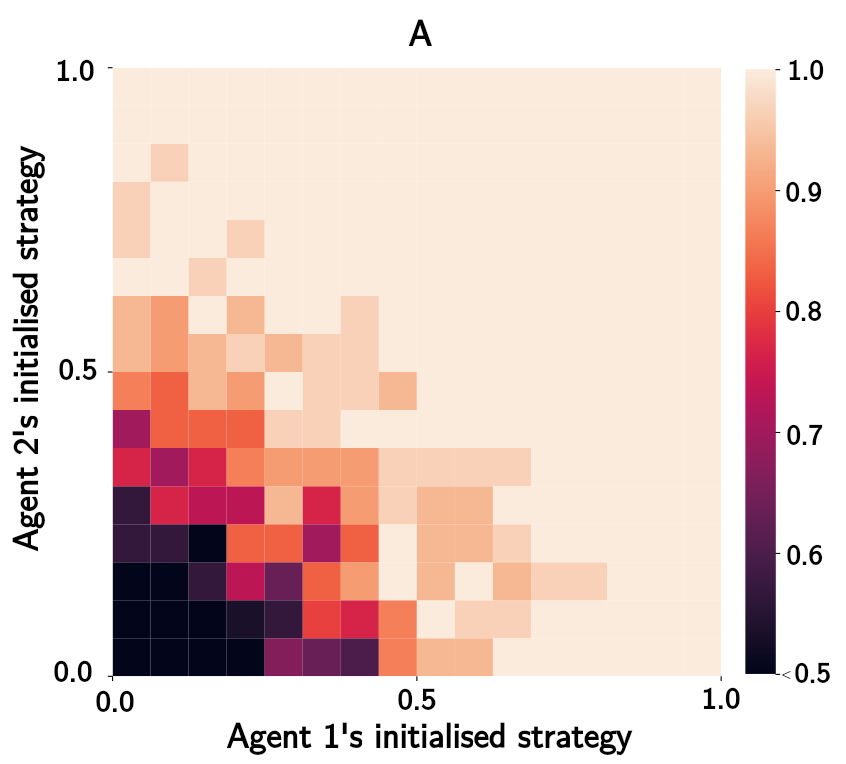}~~~~\includegraphics[width=0.4\textwidth]{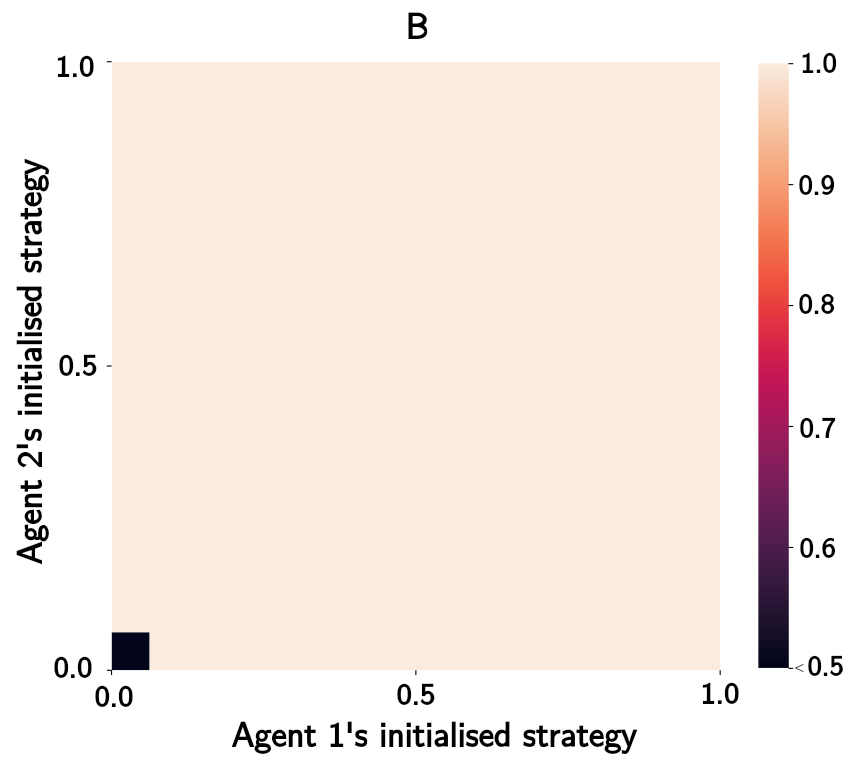}
\par\end{centering}

\caption{\label{fig:ToMAGAsCompareColorCode} Initial probability of the second
player following cooperative strategy (y-axis) vs Initial probability
of the first player following cooperative strategy (x-axis). The
colour shows the probability (lighter values indicate higher probability)
of the first player following cooperative strategy after $500$ timesteps
of (A) Guilt averse agents without theory of mind and (B) ToMAGAs.
 }
\end{figure*}
 We test our ToMAGAs in three environments: (1) Matrix-form Stag
Hunt Games; (2) Grid-world Stag Hunt Games; and (2) The modified version
of Stag Hunt Games called \emph{Island }with the more complex reward
structure and action space.

\subsection{Matrix-Form Stag Hunt Games }

\emph{}In this experiment, we aim to answer two questions:

\textbf{Q1: }\emph{How does ToM model affect cooperative behaviour
in the self-play setting?} We compare the behaviour of ToMAGAs and
GA agents \emph{without} ToM that do not update first order beliefs.
All agents have the guilt sensitivity $\theta_{ij}=200$. The initial
probabilities of each agents to follow a cooperative strategy constitute
the grid index in Figure ~\ref{fig:ToMAGAsCompareColorCode}). We
measure the probability of the agents following cooperative policy
after $500$ timesteps of playing the matrix-form games with $h=40,c=30,m=20,g=0$.
  Figure ~\ref{fig:ToMAGAsCompareColorCode} shows that ToMAGAs
promote cooperation better than the guilt averse agents without theory
of mind. This is more pronounced in settings where agents are initialised
with a low probability of following cooperative strategy (to the left
bottom corner of Figure~\ref{fig:ToMAGAsCompareColorCode}-A and
\ref{fig:ToMAGAsCompareColorCode}-B).

\textbf{Q2:} \emph{How does ToMAGA promote cooperative behaviour
in a group of agents?} The experiments are designed similarly to the
tournament commonly used in studies of how cooperation could evolve
\cite{axelrod1981evolution}. In each round, agents are randomly
matched and each pair plays the matrix-form of Stag Hunt game with
$h=5,c=4,m=2,g=1$. We report the average common reward of the last
$100$ rounds after $5000$ rounds of interaction. \begin{wrapfigure}{o}{0.5\columnwidth}%
\begin{centering}
\includegraphics[width=0.5\columnwidth]{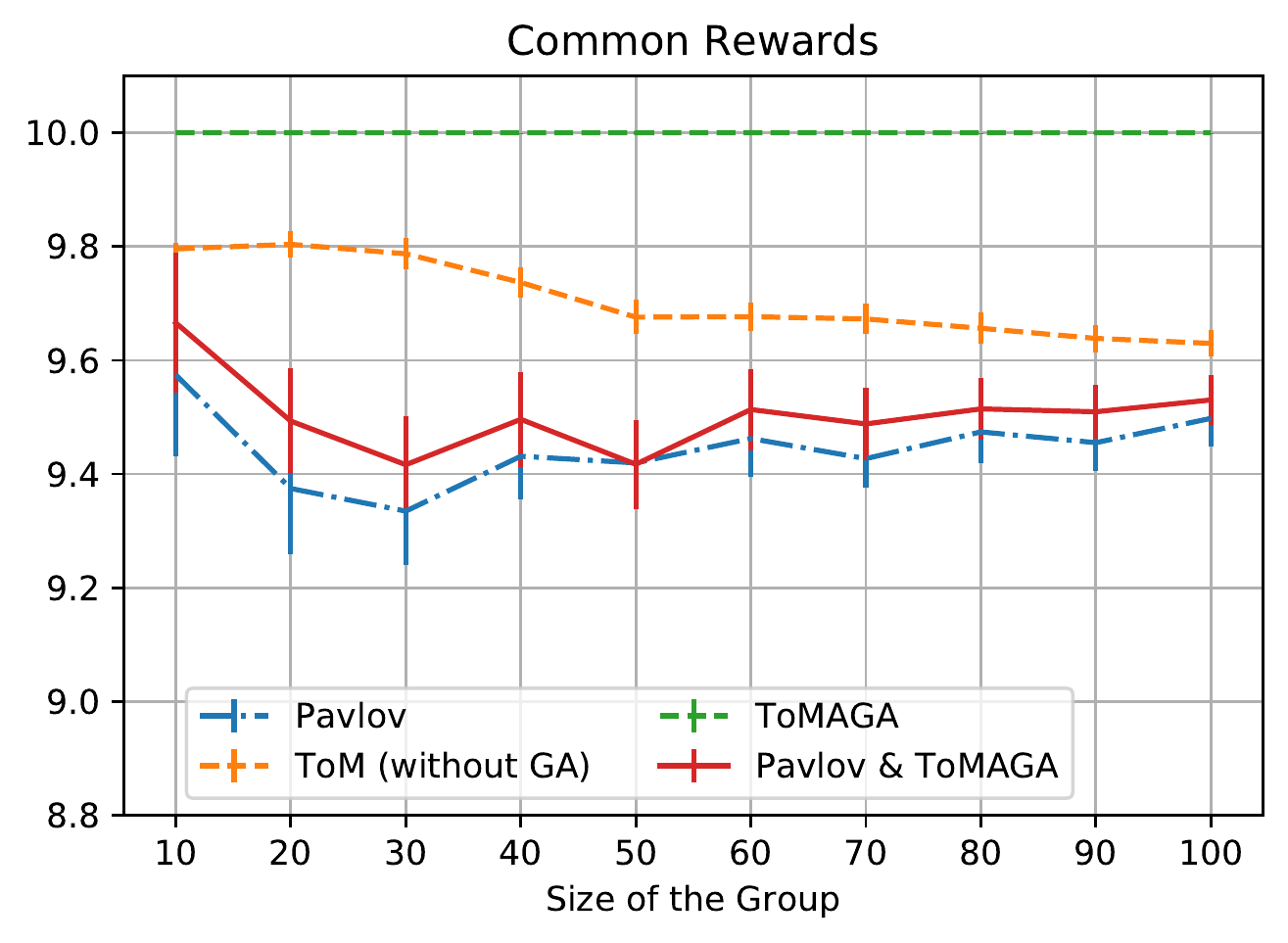}
\par\end{centering}
\caption{\label{fig:result_matrix_group-2} Common Reward (y-axis) vs Size
of Group (x-axis). ToMAGAs encourage the cooperation in both homogeneous
and heterogeneous groups. Common reward is higher when a group contains
ToMAGAs.}
\end{wrapfigure}%
 There are two types of groups: homogeneous group and heterogeneous
group; and two types of agents: ToMAGA and Pavlov agent. We
compare behaviours of ToMAGAs in groups with a general version of
Win-Stay-Lose-Shift (WSLS) strategy, called Pavlov agent, which is
a popular strategy for solving Stag Hunt. A Pavlov agent \cite{KrainesK97}
chooses to hunt stag with probability $p_{n}=\frac{i}{n}$ with $0\leq i\leq n$
and updates the strategy based on the outcome it received and actions
that both took in the last interaction. The probability cooperatively
hunt stag $p_{n}$ is increased when two players matched their behaviours,
and $p_{n}$ is decreased otherwise. In the heterogeneous group of
$N$ agents, there are $(N-1)$ Pavlov agents and a ToMAGA. From the
structure of Stag Hunt games, if one group has more agents with cooperative
behaviours, they will obtain higher average common rewards. Fig.~\ref{fig:result_matrix_group-2}
shows that the homogeneous group of ToMAGAs cooperate better than
the homogeneous group of Pavlov agents. As the size of heterogeneous
groups is small, having one ToMAGA will enhance the cooperation and
help to obtain higher common rewards. When the size of the group increases,
the homogeneous group of ToM agents without guilt aversion converge
much slower than the group of ToMAGA. This leads to the homogeneous
group of ToM agents without guilt aversion has lower common rewards
than the homogeneous group of ToM agents after $5000$ rounds of interaction.

\subsection{Grid-World Stag Hunt Games }

\begin{wrapfigure}{r}{0.5\columnwidth}%
\vspace{-6mm}

\begin{centering}
\includegraphics[width=0.5\columnwidth]{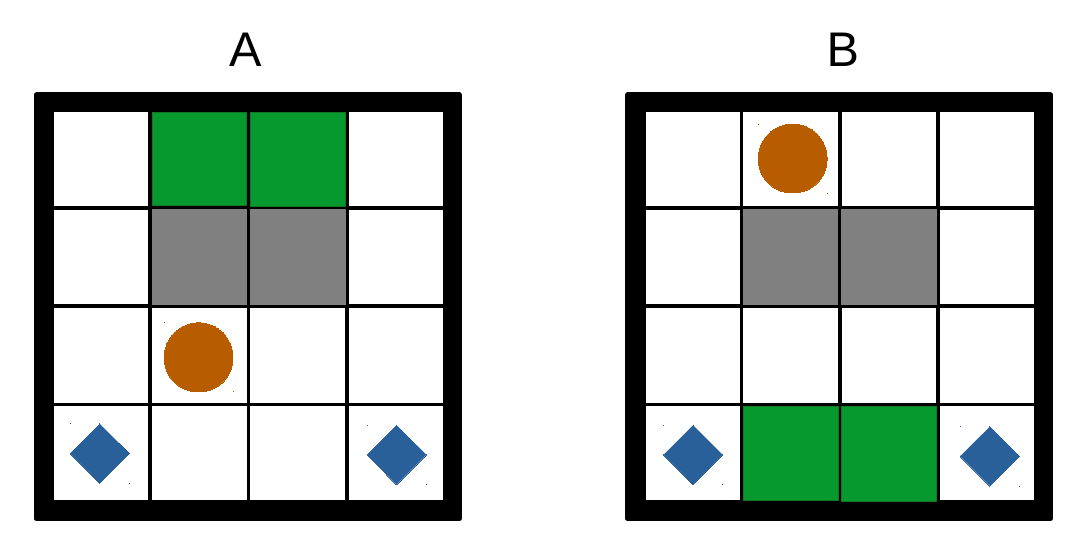}
\par\end{centering}
\caption{\label{fig:grid-settings}The grid-world version of Stag Hunt Games.
Two agents (blue diamonds) learn to hunt the moving stag (brown circle)
or the static hares (green cell) while avoiding the obstacles (in
gray). Agents start (A) start nearby the stag, and (B) nearby the
hares.}

\vspace{2mm}
\end{wrapfigure}

In the grid-world Stag Hunt games, two players simultaneously move
in a fully observable $4\times4$ grid-world, and try to catch stag
or hare by moving into their squares (see Fig.~\ref{fig:grid-settings}).
 Every timestep, each player can choose among $5$ actions $\{$\texttt{left,
up, down, right, stay}$\}$. While the players need to cooperate to
catch the stag, i.e. both move to the position of the stag at the
same time, each player can decide to catch the hare alone. The rewards
given to agents follow the reward structure of the Stag Hunt games.
In detail, if two players catch the stag together, the reward given
to each player is $4.0$. If two players catch the hares at the same
time, the reward given to each player is $2.0$. Otherwise, the player
catching the hare alone will receive a reward of $3.0$, and the other
will receive $0.0$. The game is terminated when at least one player
reaches the hare, two players catch the stag, or the time $T_{max}$
runs out. 

\begin{figure*}
\begin{centering}
\includegraphics[width=1\textwidth]{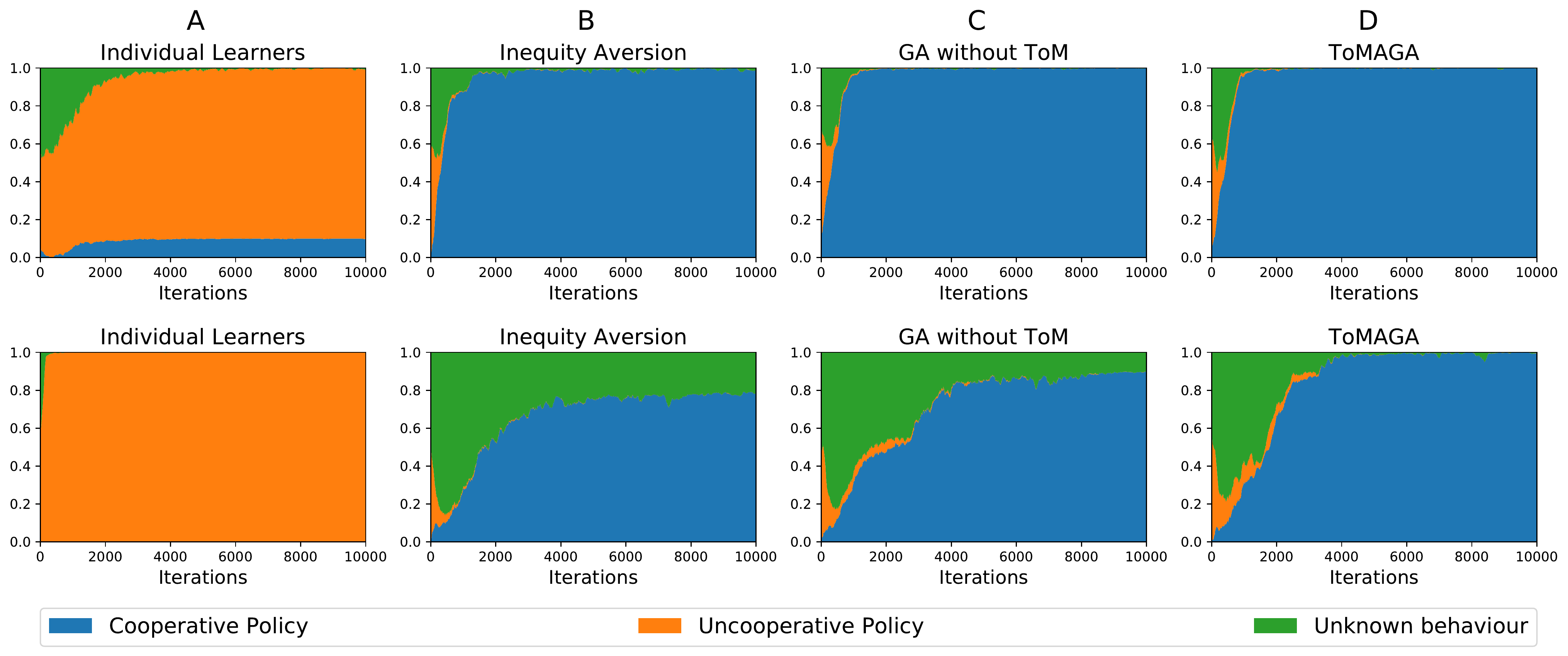}
\par\end{centering}
\caption{\label{fig:DeepRLLabels} Policies of individual learners (column
A), agents with inequity aversion (column B), GA agents without ToM
(column C), and ToMAGAs (column D) when they start nearby the stag
(the first row) and nearby hares (the second row). Proportion of following
cooperative (blue), uncooperative (orange), unknown (green) behaviours
(y-axis) vs Iterations (x-axis).}
\end{figure*}

Recall in the section \ref{subsec:The-Stag-Hunt} that both players
cooperatively catching the stag result in policies that are Pareto
efficient. We are interested in two situations: At the beginning
of the training process, agents are put (A) nearby the stag and far
from the hares, and (B) put nearby the hares and far from the stag.
We hypothesise that it is easier for agents to learn to cooperatively
catch the stag if they are put nearby the stag at the beginning. After
each iteration, each policy will be labelled as follows: (1) when
both hunt the stag, labels are $(\tilde{l}_{i}=C,\tilde{l}_{j}=C)$;
(2) when one hunts hare and other hunts stag, labels are $U$ and
$C$, respectively; (3) when both hunt hare, labels are $U$; and
(4) if the game is terminated, the policies of the agent who does
not hunt hare or stag will be considered as unknown behaviours.

We construct deep reinforcement learning agents having both value
network and policy network trained by PPO \cite{schulman2017proximal}.
We compare the behaviours of four types of agents: (1) the individual
learners; (2) the agents with inequity aversion (IA) \cite{hughes2018inequity};
(3) the GA agents without ToM; and (4) the ToMAGAs. Individual learners
are agents that behave self-interest and only optimise their rewards.
 Inequity averse agents are agents that have a shaping reward $r_{i}^{(psy)}=-\frac{\theta_{ad}}{N-1}\times{\displaystyle \sum_{j\neq i}}\max(r_{i}-r_{j},0)-\frac{\theta_{dis\_ad}}{N-1}\times{\displaystyle \sum_{j\neq i}}\max(r_{j}-r_{i},0)$,
where $N$ is the number of agents, $\theta_{ad}$ and $\theta_{dis\_ad}$
are advantageous and disadvantageous sensitivity, respectively \cite{fehr1999theory}.

Fig.~\ref{fig:DeepRLLabels} shows the policies of deep reinforcement
learning agents over the training process when they start nearby the
stag (the first row of Fig.~\ref{fig:DeepRLLabels}) and nearby the
hares (the second row of Fig.~\ref{fig:DeepRLLabels}). In both cases,
the individual learners i.e. deep reinforcement learning agents without
social preferences cannot learn to cooperate and even learn the uncooperative
behaviours (to individually catch hares) since the very early stage
of training process if they start nearby hares. In contrast, the deep
reinforcement learning agents with social preferences can learn to
cooperate in both cases. When the agents are put nearby stag at the
beginning, the performance of inequity averse agents is comparable
to GA agents without ToM and the ToMAGA (the first row of columns
B, C, and D of Fig.~\ref{fig:DeepRLLabels}). However, when initialised
nearby hares, GA agents without ToM and ToMAGA learn to cooperate
faster than the inequity averse agents (the second row of columns
B, C, and D of Fig.~\ref{fig:DeepRLLabels}). Also, in this case,
our ToMAGAs can learn to cooperate faster than the GA without ToM
because the GA without ToM does not update its first-order belief,
leading to wrong predictions about the expectation of others.

\subsection{The Island: ToMAGA in a Complex Environment}

\begin{figure*}
\begin{centering}
\includegraphics[width=1\textwidth]{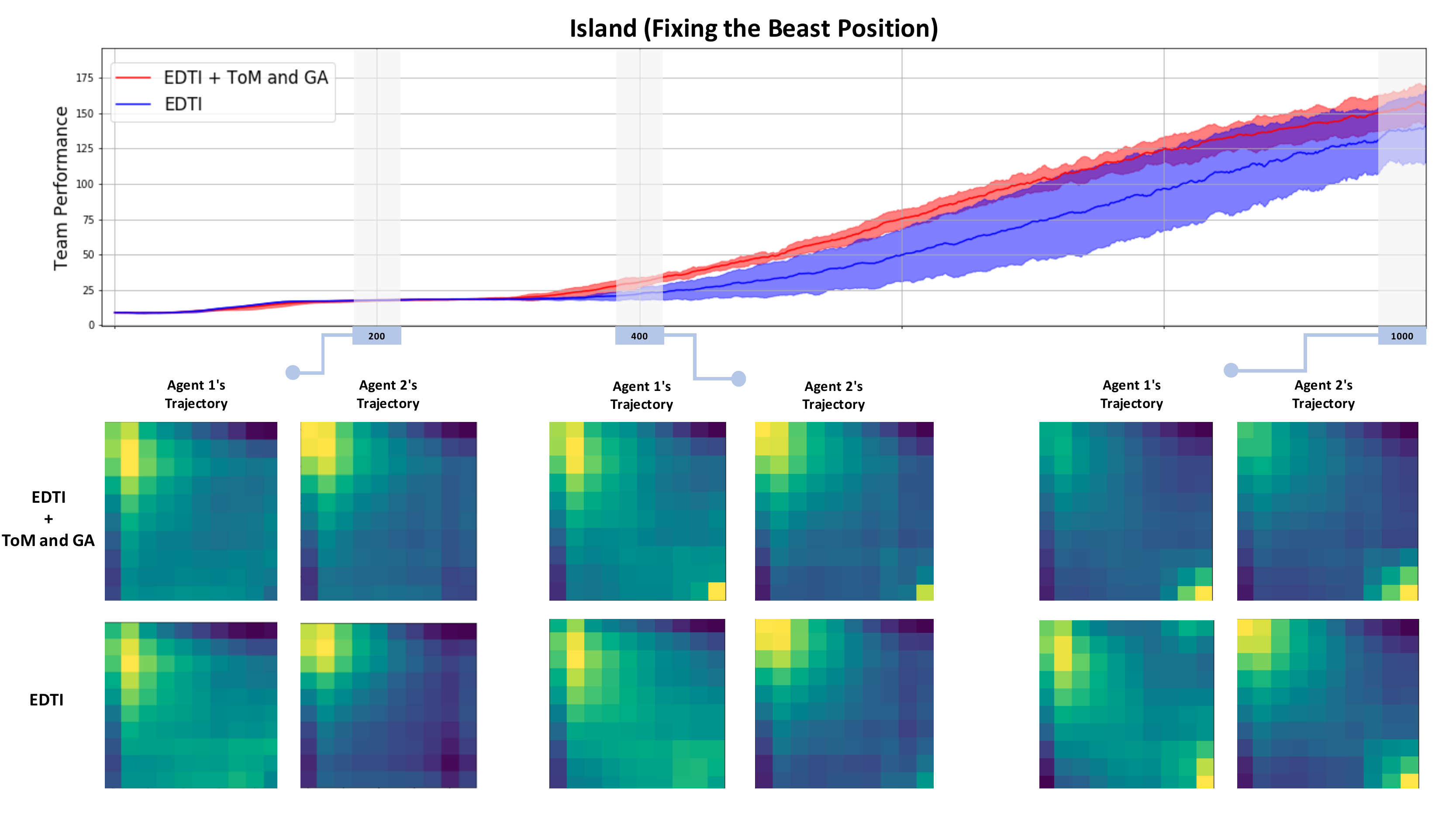}
\par\end{centering}
\caption{\label{fig:IllustrationIslandFixedBeast} The behaviours of the EDTI
agents and EDTI agents augmented with theory of mind and guilt aversion
in the $10\times10$ Island environment with static beast. The upper
part is Team Performance (y-axis) vs Number of Updates (x-axis). The
lower part is the visitation count of the agents overtime. A cell
with lighter colour means that it is visited more frequently by the
agents. EDTI agents augmented with theory of mind and guilt aversion
learns to cooperative attack the beast faster and tend to maintain
equity in the team.}
\end{figure*}
\begin{figure*}
\begin{centering}
\includegraphics[width=0.5\textwidth]{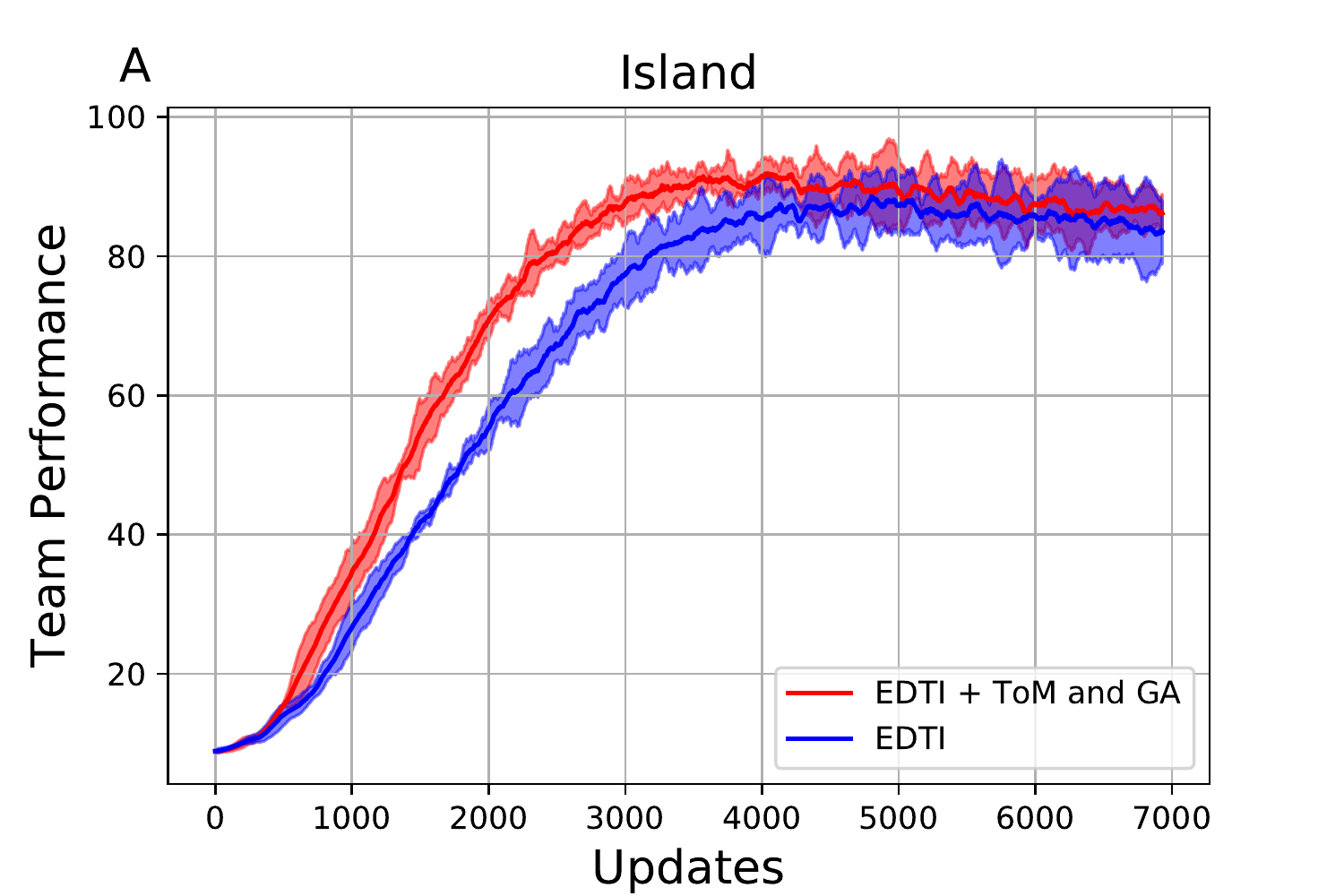}\includegraphics[width=0.5\textwidth]{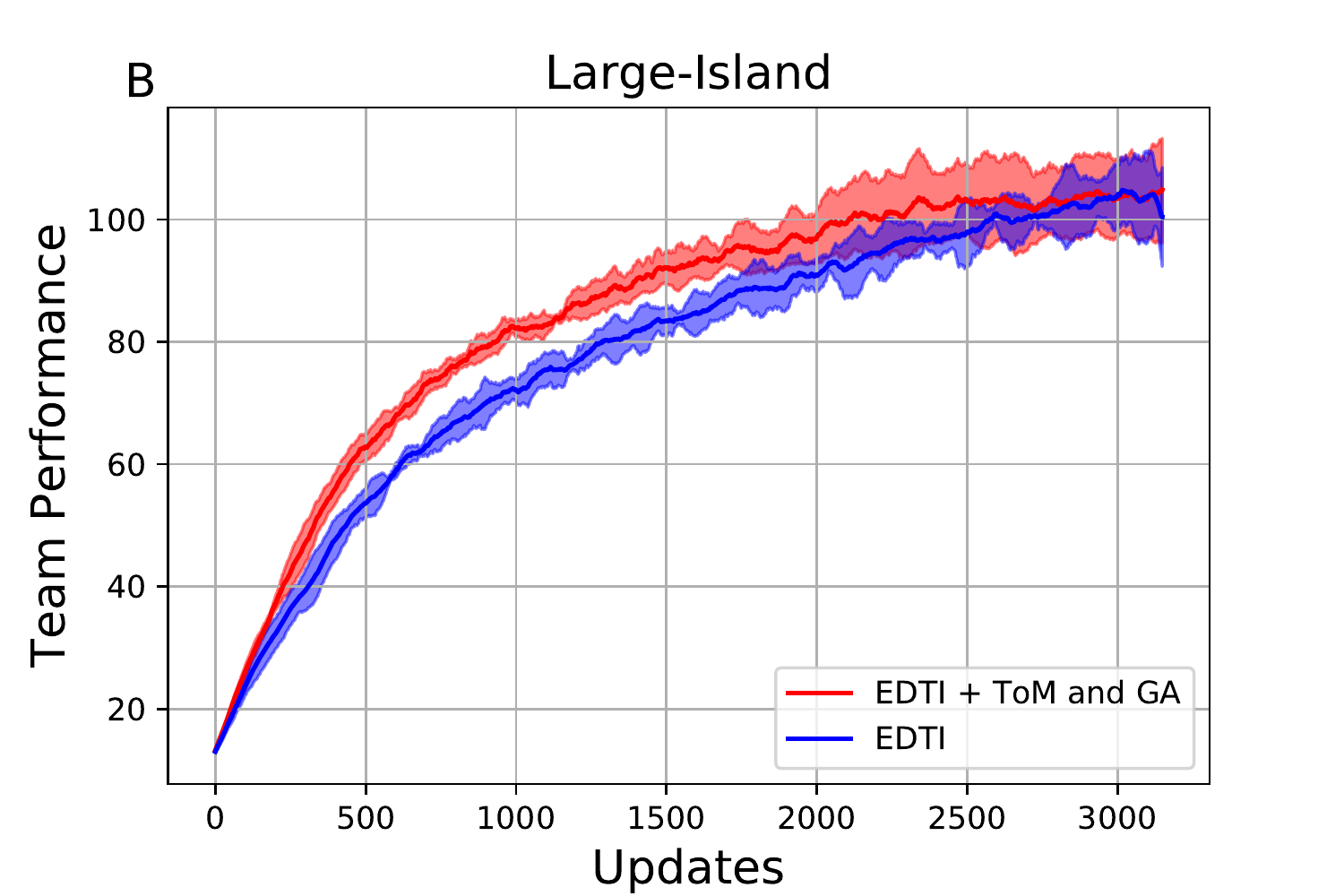}
\par\end{centering}
\caption{\label{fig:IslandReward} Team Performance (y-axis) vs Number of Updates
(x-axis). The performance of EDTI agents and EDTI agents augmented
with theory of mind and guilt aversion on complex environments, which
are: (A) Island and (B) Large-Island.}
\end{figure*}

This suite of experiments aims to demonstrate the performance of our
reward shaping mechanism on more complex environments, we consider
the behaviours of ToMAGA on the Island and its extended version with
more agents (Large-Island) \cite{wang2020edti}. These environments
are modified versions of the Stag Hunt game with more complex rewards
structure and action space. In Island game, there are two agents and
a beast in a $10\times10$ environment. Instead of only moving around
by choosing \texttt{left, up, down, right, stay} as in the grid-world
Stag Hunt games, the agent must get close and choose \texttt{attack}
to kill the beast. Similarly, the beast also can move and attack the
agents which are in its attack range. The beast and agents will have
its own energy which will be reduced if they are attacked, and one
will be killed if their energy is equal $0$. The agents will receive
a reward $300$ if they kill the beast. In this island, there are
treasures, which an agent only individually receives a reward of $10$
if they collect. If the agents jointly attack the beast, they will
kill the beast faster and reduce the chance of being killed. In the
Large-Island, we aim to test our reward shaping mechanism with more
than two agents. In this environment, the team with four agents will
explore the $4\times4$ island.

We build our rewards shaping mechanism to the agent Exploration via
Decision-Theoretic Influence with intrinsic rewards (EDTI) \cite{wang2020edti}.
In both environments, the labels of agents' policies will be given
to agents at the end of each episode. An agent will be considered
as following cooperative behaviour if this agent attacks the beast
at least once during the episode. In the Large-Island, to handle the
interaction of more than two agents, an agent $i$ will have the first-order
belief not only about what other believes about itself, but about
what other believes about others. Therefore, the first-order beliefs
of agents are implemented as described in \cite{burkhard2017manyminds}.
For example, the first-order belief of agent $i$ is $b_{i}^{(1)}=\left\{ b_{j,k}^{(1)}|j\in\mathcal{N}\setminus\{i\},k\in\mathcal{N}\right\} $,
which is what agent $i$ believes about agent $j$ believes about
agent $k$ for all $j\in\mathcal{N}\setminus\{i\},k\in\mathcal{N}$.

To illustrate the behaviours of the EDTI agent and EDTI agent augmented
with theory of mind and guilt aversion, we fix the beast on the Island
at the cell $(9^{\text{th}},9^{\text{th}})$ (on the right bottom
corner of the environment) and the treasures at cells $(1^{\text{th}},1^{\text{th}})$
and $(2^{\text{th}},1^{\text{th}})$ (on the left top corner of the
island). Figure \ref{fig:IllustrationIslandFixedBeast} shown the
team performance which is the average of team rewards and the visitation
of agents (cells that have ligher colour are cells that visited more
frequently by agents). In the earlier phase, both type of agents tend
to visit the top left of the environment and collect the treasures.
Over time, the agents start to discover the position of the beast
and learn how to attack the beast. The EDTI agents augmented with
theory of mind discovered the strategy of together attacking the beast
and focus on visiting cells nearby the beast faster than the EDTI
agents. It worth noting that the EDTI agent $1$ finds the beast and
attacks the beast before the EDTI agent $2$, .i.e. the right bottom
corner of the visitation map of the agent $1$ is lighter than the
visitation map of the agent $2$, which shows the phenomena of inequity
between two EDTI agents in this particular setting. In contrast,
because the EDTI agents augmented with theory of mind and guilt aversion
try to match the expectation of each other, they tend to preserve
the equity. The team performance in the Island with moving beast
is shown in Figure \ref{fig:IslandReward}-A. Figure \ref{fig:IslandReward}-B
shown the efficient of augmenting EDTI with theory of mind and guilt
aversion in the Large-Island. This demonstrates that our rewards shaping
mechanism can be extended to the setting with more than two agents.
% \end{document}

\section{Related Works}
% %% LyX 2.3.5.2 created this file.  For more info, see http://www.lyx.org/.
% %% Do not edit unless you really know what you are doing.
% \documentclass[british]{jmlr}
% \usepackage[T1]{fontenc}
% \usepackage[latin9]{inputenc}
% \usepackage{amstext}
% \usepackage{babel}
% \begin{document}
Theory of Mind \cite{gopnik1992child,premack1978does,gordon1986folk},
or the ability of understanding that other having mental states, is
crucial ability for an agent which involves in social interactions.
In economics, it is studied as forecasting the forecast of other \cite{townsend1983forecasting}.
In multi-agent system, it is known as modelling others \cite{albrecht2018autonomous}
to reduce the non-stationary problem while learning agents are updating
their models simultaneously. Recent works in cognitive science and
artificial intelligence have proposed several computational model
of Theory of Mind such as the Bayesian Theory of Mind \cite{baker2011bayesian,baker2017rational,yoshida2008game}
and Machine Theory of Mind \cite{rabinowitz2018machine}. In \cite{de2013much},
authors used $\text{ARIMA(0,1,1)}$ to model theory of mind as the
recursive reasoning about other. The agent with Theory of Mind level-1
will hold a belief about what other agents think about its action.
\cite{burkhard2017manyminds} extended this model to the settings
in which there are more than two agents, which requires each agent
holds belief about the believe of other not only about itself but
also about others, i.e. thinking about other thinking about other.
We used the Theory of Mind level-1 models described in \cite{de2013much,burkhard2017manyminds}
for belief about the higher level of actions to construct the belief
based guilt aversion agent.

Solving social dilemma still is a challenge for reinforcement learning
agents \cite{peysakhovich2018towards}. Using behavioural game theory
as prior knowledge, recent works demonstrated that inequity averse
agents \cite{hughes2018inequity} and prosocial agents \cite{peysakhovich2018prosocial}
can promote cooperation in social dilemma. However, belief based guilt
aversion \cite{battigalli2007guilt}, which is a well known mechanism
in psychological game theory to promote cooperation, has not been
studied in multi-agent reinforcement learning. \cite{moniz2017social,rosenstock2018s}
only considered the guilt without theory of mind model agents on evolutionary
dynamics. We instead use the guilt aversion with theory of mind model
to shape the reward of reinforcement learning agents.
% \end{document}

\section{Conclusion}
% %% LyX 2.3.5.2 created this file.  For more info, see http://www.lyx.org/.
% %% Do not edit unless you really know what you are doing.
% \documentclass[british]{jmlr}
% \usepackage[T1]{fontenc}
% \usepackage[latin9]{inputenc}
% \usepackage{babel}
% \begin{document}
We present a new emotion-driven multi-agent reinforcement learning
framework in which reinforcement learning agents are equipped with
theory of mind and guilt aversion - the emotion faculty that induces
a utility loss in an agent if it believes that its action has caused
harm in others. We studied the agent behaviours in Stag Hunt games,
which simulate social dilemmas, whose Pareto optimal equilibrium demands
cooperation between agents making it hard for pure reinforcement learning
agents. We validated the framework in three environments for Stag
Hunt games. Our results demonstrate the effectiveness of belief-based
guilt aversion over other methods.
% \end{document}

\bibliography{Tom2Psy}

\end{document}